\documentclass{article}
\pdfoutput=1

\usepackage[nonatbib, final]{neurips_2018}

\usepackage[utf8]{inputenc} % allow utf-8 input
\usepackage[T1]{fontenc}    % use 8-bit T1 fonts
\usepackage{hyperref}       % hyperlinks
\usepackage{url}            % simple URL typesetting
\usepackage{booktabs}       % professional-quality tables
\usepackage{amsfonts}       % blackboard math symbols
\usepackage{nicefrac}       % compact symbols for 1/2, etc.
\usepackage{microtype}      % microtypography
\usepackage{mathptmx} % This is Times font
\usepackage{graphicx,subcaption}
\usepackage{fancyhdr}
\usepackage[normalem]{ulem}
\usepackage{hyperref}
\usepackage{amsfonts,amssymb,amsmath}       % blackboard math symbols
\usepackage[table,xcdraw]{xcolor}
\usepackage{algorithmic}
\usepackage{multirow}
\usepackage{comment}
\usepackage{tikz}
\usepackage{xspace}
\usepackage{mathptmx}
\usepackage{graphicx,subcaption}
\usepackage[flushleft]{threeparttable}
\usepackage{algorithm2e, setspace}
\usepackage{enumitem}
\usepackage{marginnote}
\usepackage{float}
\usepackage{authblk}
\usepackage{wrapfig}
\usepackage{setspace}
\usepackage[export]{adjustbox}

\iftrue
\newcommand{\new}[1]{{\color{blue}#1}}
\newcommand{\alex}[1]{{\color{orange}[\textbf{\sc alex}: \textit{#1}]}}
\newcommand{\youjie}[1]{{\color{violet}[\textbf{\sc youjie}: \textit{#1}]}}
\else
\newcommand{\new}[1]{}
\newcommand{\alex}[1]{}
\newcommand{\youjie}[1]{}
\fi
\makeatletter
\def\@onedot{\ifx\@let@token.\else.\null\fi\xspace}
\DeclareRobustCommand\onedot{\futurelet\@let@token\@onedot}

\makeatletter
\renewcommand{\@algocf@capt@plain}{above}
\makeatother
\def\eg{\textit{e.g}\onedot} 
\def\ie{\textit{i.e}\onedot} 
 
\def\etc{\textit{etc}\onedot} \def\vs{\textit{vs}\onedot}
\def\wrt{w.r.t\onedot} 
\def\etal{\textit{et al}\onedot}
\def\OURS{\textsf{\footnotesize Pipe-SGD}\xspace}

\newcommand{\tline}[0]{\vspace{2pt}\hrule\vspace{2pt}}
\newcommand{\bline}[0]{\vspace{2pt}\hrule\vspace{2pt}}

\newcommand{\beas}{\begin{eqnarray*}}
\newcommand{\eeas}{\end{eqnarray*}}
\newcommand{\bea}{\begin{eqnarray}}
\newcommand{\eea}{\end{eqnarray}}
\newcommand{\bes}{\begin{equation*}}
\newcommand{\ees}{\end{equation*}}
\newcommand{\be}{\begin{equation}}
\newcommand{\ee}{\end{equation}}

\newcommand{\cD}{{\cal D}}

\newcommand{\cB}{{\cal B}}

\newcommand{\figref}[1]{Fig\onedot~\ref{#1}}
\newcommand{\equref}[1]{Eq\onedot~\eqref{#1}}
\newcommand{\secref}[1]{Sec\onedot~\ref{#1}}

\newcommand{\algref}[1]{Alg\onedot~\ref{#1}}

\title{Pipe-SGD: A Decentralized Pipelined SGD Framework for Distributed Deep Net Training}

\author[$\dag$]{\textbf{Youjie Li}} %% li238@illinois.edu
\author[*]{\textbf{Mingchao Yu}} %% fishermanymc@gmail.com
\author[*]{\textbf{Songze Li}} %% songzeli@usc.edu
\author[*]{\textbf{Salman Avestimehr}} %% avestimehr@gmail.com
\author[$\dag$]{\authorcr\textbf{Nam Sung Kim}} %% nam.sung.kim@gmail.com
\author[$\dag$]{\textbf{Alexander Schwing}} %% aschwing@illinois.edu

\affil[$\dag$]{University of Illinois at Urbana-Champaign}
\affil[*]{University of Southern California}
\begin{document}

\maketitle

%%%%%%% -- PAPER CONTENT BEGIN -- %%%%%%%%
%\input{abstract} -------------------------------------------------------------------------------------------
\begin{abstract}
Distributed training of deep nets is an important technique  to address some of the present day computing challenges like memory consumption and computational demands. %helping to also address bottlenecks like memory consumption. 
%% NSK
%Especially, exchanging a large amount of parameters and gradients among compute nodes every iteration over the network becomes the critical bottleneck to train a large neural net fast. 
%% NSK
Classical distributed approaches, synchronous or asynchronous, are  based on the parameter server architecture, \ie, worker nodes compute gradients which are communicated to the parameter server while  updated parameters are returned. Recently, distributed training with  AllReduce operations gained popularity as well. %Upon updating the parameters synchronously or asynchronously, neural net weights are sent back to the workers. 
%To improve applicability of compression techniques for faster communication, in this paper, we propose a novel communication scheme based on gradients only. In experiments on a variety of benchmark datasets we obtain a significant reduction in communication time. For smaller models this improvement is mitigated to some degree by the time for compression.
While many of those operations seem appealing, little is reported about wall-clock training time improvements.  In this paper, we carefully analyze the AllReduce based setup, propose timing models which include network latency, bandwidth, cluster size and compute time, and demonstrate that a pipelined training with a width of two combines the best of both synchronous and asynchronous training. Specifically, for a setup consisting of a four-node GPU cluster we show wall-clock time training improvements of up to $5.4\times$ compared to conventional approaches. %without degradation in accuracy.
\end{abstract}
\section{Introduction}
\label{sec:introduction}

Deep nets~\cite{LeCunNature2015,BengioPAMI2013} are omnipresent across  fields from computer vision and natural language processing to computational biology and robotics. Across  domains and tasks they have demonstrated impressive results by automatically extracting hierarchical abstractions of representations from many different datasets. The surge in popularity pivoted in the 2010s, with impressive results being demonstrated on the ImageNet dataset~\cite{KrizhevskyNIPS2012,RussakovskyIJCV2015}. 
Since then, deep nets have been applied to many more tasks. Prominent examples include recognition of places~\cite{ZhouNIPS2014}, playing of Atari games~\cite{MnihNIPSWS2013,MnihNature2015}, and the game of Go~\cite{SilverNature2016}. %, and more recently dermatologist-level classification of skin cancer~\cite{EstevaNature2017}. 
Common to all those methods is the use of large datasets to fuel the many layers of deep nets. 

Importantly, in the last few years, the number of layers, or more generally the depth of the computation tree has increased significantly from a few layers for LeNet~\cite{LeCunIEEE1998} to several 100s or 1000s~\cite{HeCVPR2016,LarssonARXIV2016}. Inherent to the increasing complexity of the computation graph is an increase in training time and often also an increase in the amount of data that is processed. Traditionally, computational performance increases do not keep up with the desired processing needs despite the use of accelerators like GPUs. 
%% NSK - begin
%Importantly, parallelization and sample parallel distributed training provides a compelling framework to easily increase processing capabilities, albeit at the expense of communication efforts, which often have a significant impact.
%% what do you mean by ``sample parallel''?
%Therefore, parallelization and/or acceleration with Graphic Processing Unit (GPU), Field Programmable Gate Array (FPGA) and Application-Specific Integrated Circuit (ASIC) provides a compelling framework to increase processing capabilities.

Beyond accelerators, parallelization of computation on multiple computers is therefore popular. However, it  requires frequent communication to exchange  a large amount of data among compute nodes while the bandwidth of network interfaces is limited. 
This in turn significantly diminishes the benefit of parallelization, as a substantial fraction of training time is spent to communicate data. 
The fraction of time spent on communication is further increased when applying accelerators~\cite{IandolaCVPR2016,CMU_EuroSys16,Nvidia15, Cong_FPGA15,Li_ISCAS16,Li_NeuroCom17,dnnweaver}, as they  decrease computation time while leaving communication time untouched. 
%---------------------------------------
%\begin{figure}[!t]
%  \centering
%  \includegraphics[width=0.5\linewidth]{figures/modelsize-comptime.pdf}
%  \caption{(a) The size of weights (or gradients). (b) The percentage of the time spent to exchange $g$ and $w$ in total training time with a conventional worker-aggregator approach.} 
%  \youjie{Need this figure in introduction?}
%%  \vspace{-3ex}
%  \label{fig:size_time}
%\end{figure}
%----------------------------------------

To take advantage of parallelization across machines, a variety of approaches have been developed starting from the popular MapReduce paradigm~\cite{DeanACM2008,ZahariaUSENIX2010,Isard2007,Murray2013}. Despite their benefits, communication heavy training of deep nets is often based on custom implementations~\cite{Dean2012,Chilimbi2014,MoritzICLR2016,Kim2016} relying on the parameter server architecture~\cite{MuLiNIPS2014, MuLiOSDI2014, SSP}, where the centralized server aggregates the gradients from workers and distributes the updated weights, either in a synchronous or asynchronous manner. Recent research proposed to use a decentralized architecture with global synchronization among nodes~\cite{Facebook1Hour, deep_g_compression}. However, in common to all the aforementioned techniques, little is reported regarding the timing analysis of distributed deep net training. 

%Importantly, in common to all the aforementioned parameter server techniques is the transfer of gradients to a parameter server which maintains and distributes the accumulated weights, either synchronously or asynchronously to the worker nodes.

In this paper, we  analyze the wall-clock time trade-offs between communication and computation. To this end we develop a model to assess the training time based on a set of parameters such as latency, cluster size, network bandwith, model size, \etc. Based on the results of our model we develop \OURS, a framework with pipelined training and balanced communication, and show its convergence properties by adjusting proofs of~\cite{langford2009slow, SSP}. We also show what types of compression can be efficiently included in an AllReduce based framework. Finally, we assess the speedups of our proposed approach on a GPU cluster of four nodes with 10GbE network, showing wall-clock time training improvements by a factor of $3.2\sim5.4\times$ compared to conventional centralized and decentralized approaches
%when using {\color{blue}xxx} 
without degradation in accuracy.
\section{Background}
\label{sec:background}

\textbf{General Training of Deep Nets:} %Efficient optimization in general and training of deep nets in particular is an important problem. 
%Training of deep nets involves finding the parameters $w$ of a predictor $F(x,w)$ given input data $x$.
%In supervised learning scenario, we choose the parameters $w$ of the predictor $F(x,w)$ by minimizing a 
%loss function $\ell(F(x,w),y)$ which compares the predictor output $F(x,w)$ for given data $x$ and the current $w$ to the ground-truth annotation $y$. Given such a dataset $\cD = \{(x,y)\}$,  the process of finding $w$ is formally summarized in the following objective:
%\be
%\min_w f_\cD(w) := R(w) + \frac{1}{|\cD|}\sum_{(x,y)\in\cD} \ell(F(x,w),y),
%\label{eq:Obj}
%\ee
%where $R(w)$ denotes a regularization term.
%
%Optimization of the objective given in \equref{eq:Obj} \wrt the parameters $w$, \eg, via gradient descent using $\frac{\partial f_\cD}{\partial w}$, can be challenging due not only to complexity of  evaluating the predictor $F(x,w)$ and its derivative but also to the size of the dataset $|\cD|$. Consequently, stochastic gradient descent (SGD) emerged as a popular technique. We randomly sample a subset $\cB$ of the dataset, often also referred to as a minibatch. Instead of computing the gradient on the entire dataset $\cD$, we approximate it using the samples in the minibatch, \ie, we assume $\frac{\partial f_\cD}{\partial w} \approx \frac{\partial f_\cB}{\partial w}$. However, for present day datasets and predictors, computation of the gradient $\frac{\partial f_\cB}{\partial w}$ on a single machine is still challenging. Minibatch sizes $|\cB|$ of less than 20 samples are common, \eg, when training semantic image segmentation systems~\cite{ChenICLR2015}.
%
Training of deep nets involves finding the parameters $w$ of a predictor $F(x,w)$ given input data $x$.
%Training results in parameters $w$ of the predictor $F(x,w)$ by
To this end we minimize a 
loss function $\ell(F(x,w),y)$ which compares the predictor output $F(x,w)$ for given data $x$ and the current $w$ to the ground-truth annotation $y$. Given  a dataset $\cD = \{(x,y)\}$, finding $w$ is formally summarized  via:
\be
\min_w f_\cD(w) := \frac{1}{|\cD|}\sum_{(x,y)\in\cD} \ell(F(x,w),y).
\label{eq:Obj}
\ee
%where $R(w)$ denotes a regularization term.
Optimization of the objective given in \equref{eq:Obj} \wrt the parameters $w$, \eg, via gradient descent using $\frac{\partial f_\cD}{\partial w}$, can be challenging due to not only the complexity of evaluating the predictor $F(x,w)$ and its derivative, but also the size of the dataset $|\cD|$. Consequently, stochastic gradient descent (SGD) emerged as a popular technique. We randomly sample a subset $\cB$ of the dataset, often also referred to as a minibatch. Instead of computing the gradient on the entire dataset $\cD$, we approximate it using the samples in the minibatch, \ie, we assume $\frac{\partial f_\cD}{\partial w} \approx \frac{\partial f_\cB}{\partial w}$. However, for present day datasets and predictors, computation of the gradient $\frac{\partial f_\cB}{\partial w}$ on a single machine is still challenging. Minibatch sizes $|\cB|$ of less than 20 samples are common, \eg, when training for semantic image segmentation~\cite{ChenICLR2015}.

\textbf{Distributed Training of Deep Nets:} To train larger models or to increase the minibatch size,  distributed training on multiple compute nodes is used~\cite{Dean2012,SSP,Chilimbi2014,MuLiOSDI2014,MuLiNIPS2014,MoritzICLR2016,IandolaCVPR2016}. A popular architecture to facilitate distributed training is the parameter server framework~\cite{SSP,MuLiOSDI2014,MuLiNIPS2014}. The parameter server %, as the names suggest, 
maintains a copy of the current parameters, and communicates with a group of worker nodes, each of which operates on a small minibatch to compute local gradients based on the retrieved parameters $w$. Upon having completed its task, the worker shares the gradients with the parameter server. Once the parameter server has obtained all or some of the gradients it updates the parameters using the negative gradient direction and afterwards shares the latest values with the workers. 

%------------------------------------------------------------------------------
%\begin{figure}[!tbp]
%  \begin{subfigure}[b]{0.35\textwidth}
%    \includegraphics[width=\textwidth]{figures/PS-Timeline.pdf}
%    \caption{Parameter Server with Asynchronous Training.}
%    \label{fig:PS-Async}
%  \end{subfigure}
%  \hfill
%  \begin{subfigure}[b]{0.35\textwidth}
%    \includegraphics[width=\textwidth]{figures/DSync-Timeline.pdf}
%    \caption{Decentralized Synchronous Training.}
%    \label{fig:D-Sync}
%  \end{subfigure}
%  \hfill
%  \begin{subfigure}[b]{0.25\textwidth}
%    \includegraphics[width=\textwidth]{figures/DAsync-Timeline.pdf}
%    \caption{Decentralized Pipeline Training.}
%    \label{fig:D-Pipeline}
%  \end{subfigure}
%  
%   	\caption{Comparison between different distributed learning frameworks.{\color{blue}can we align those figures a little more and maybe make them slightly bigger?}}
%	\label{fig:AllFrame}
%	\vspace{-1.5ex}
%\end{figure}

\begin{figure}[!t]
	\center
    \includegraphics[width=1.0\linewidth]{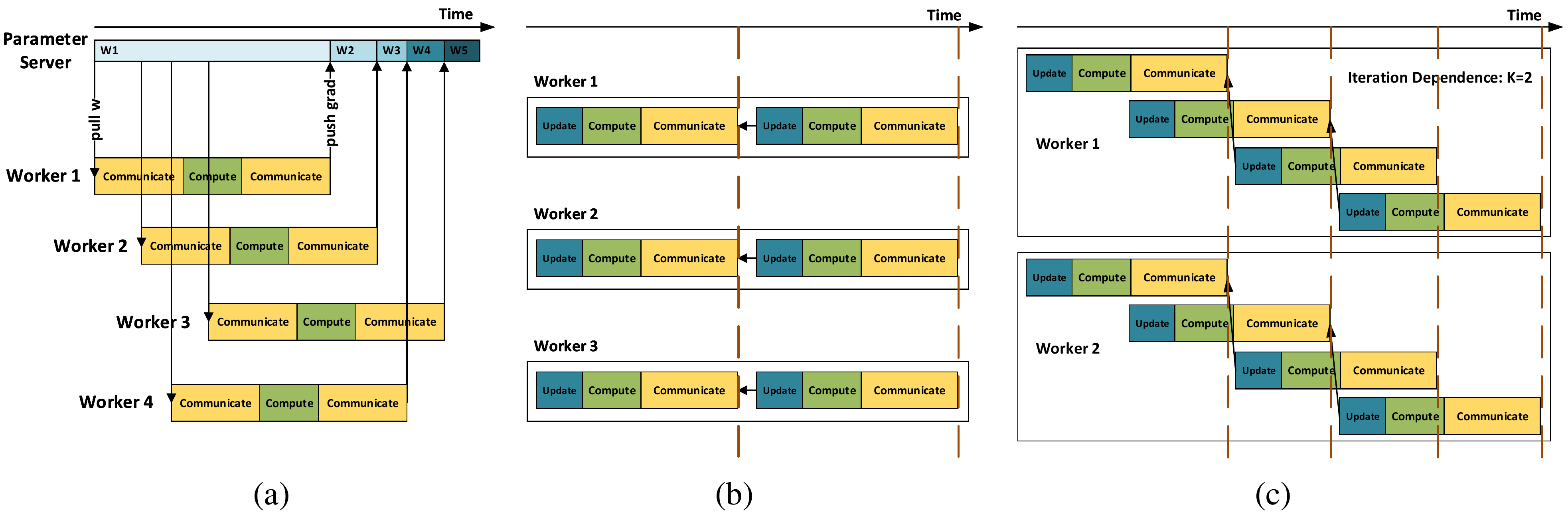}
    %\vspace{-1ex}
   	\caption{Comparison between different distributed learning frameworks: (a) parameter server with asynchronous training, (b) decentralized synchronous training, and (c) decentralized pipeline training.}
	\label{fig:AllFrame}
    \vspace{-2.5ex}
\end{figure}
%------------------------------------------------------------------------------
Asynchronous updates where each worker  independently pulls $w$ from the server, computes its own local gradient, and  pushes results back are available and illustrated in \figref{fig:AllFrame} (a). Due to the asynchrony, minimal synchronization overhead is traded with staleness of gradients. Methods for staleness control  exist, which  bound the number of  delay steps~\cite{SSP}. %The method of bounded iteration drift for workers, however, still suffers from staled update, since workers within the bound can still commit their updates to the centralized server such that iterations of stale update can be linear to the cluster size.
However, note that stale gradients %are suboptimal and 
may slow down training significantly. 

Importantly, all those frameworks are based on a centralized compute topology which  forms a  communication  bottleneck,  increasing the  training time as the cluster size scales. The time taken by pushing gradient, update, and pulling  $w$ can be linear in the cluster size due to network congestion. % In the asynchronous-update setting,  network congestion on the parameter server leads to more sever staleness. % of gradient update, as some workers might suffer from significant longer latency.

Therefore, most recently, decentralized training frameworks gained popularity in both the synchronous and asynchronous setting~\cite{DPSGD, ADPSGD}. %The former enforces a global barrier for all workers in every iteration to perform synchronous update and the latter allows workers to run at their own pace with asynchronous update. 
However, those approaches assume  decentralized workers are either completely synchronous (as in \figref{fig:AllFrame} (b)) or completely asynchronous, which requires  to either deal with long execution time every iteration or pay for uncontrolled gradient staleness. % of gradient update.  

\textbf{Compression in Distributed Training:} 
As the model size increases and cluster size scales, communication overhead in distributed learning system dominates the training time, \eg, up to $80\sim90\%$ even in a high-speed network environment~\cite{INCEPTIONN, Aluminum}. %\youjie{I think this is a good motivation number to put in paper. We can place it here or even in the introduction.} 
To reduce the communication time, various compression algorithms have been proposed recently~\cite{Microsoft_1BitSGD, Amazon_FixThresh, Lawrence_Adaptive, IBM_AdaComp, TernGradNIPS2017, deep_g_compression, QSGDNIPS2017}, some of which focus on reducing the precision of communicated gradients through scalar quantization into 1 bit, while others focus on reducing the quantity of gradients to be transferred. % by a certain magnitude threshold either constant or adaptive. 
Most compression works, however, only emphasize on achieving high compression ratio or low loss in accuracy without reporting the wall-clock training time. 

In practice,  compression without knowledge of the communication process is usually counter-productive~\cite{INCEPTIONN}, \ie, the total training time often increases. % after applying the compression, not only because compression themselves consume a substantial amount of computation time, 
%but also because the gradient communication, such Allreduce, are multi-step algorithm which requires transferred gradient to be compressed and decompressed repeatedly with a worst-case complexity linear to the cluster size. 
This is due to the fact that  AllReduce is a multi-step algorithm which requires transferred gradients to be compressed and decompressed repeatedly with a worst-case complexity linear in the cluster size, as we discuss below in~\secref{sec:compression}. 
%\new{WANT TO REMOVE: To avoid those issues in the following we briefly summarize the proposed distributed setup which was obtained from a detailed timing model.} \youjie{redundant due to the first line in overview.}
%
%\input{approach} -------------------------------------------------------------------------------------------
\section{Decentralized Pipelined Stochastic Gradient Descent}
\label{sec:app}

\textbf{Overview:} To address the aforementioned issues (network congestion for a central server, long execution time for synchronous training, and  stale gradients in asynchronous training) we propose a new decentralized learning framework, \OURS, shown in~\figref{fig:AllFrame} (c). It balances  communication  among nodes via AllReduce and pipelines the local training iterations to hide  communication time.

We developed  \OURS  by analyzing a timing model for wall-clock train time under different resource conditions using various communication approaches. We find that the proposed \OURS  is optimal when gradient updates are delayed by only one iteration and the time taken by each iteration is dominated by local computation on workers. %In this case we achieve the optimal setting of our framework via local interleaving of two neighboring iterations. 
Moreover, we found  lossy compression to further reduce communication time without impacting accuracy. %Our compression turns out be effective in evaluation on wallclock runtime, not only because of its simplicity but also because the compression benefit is maximized by the gradient-only communication in our design as compared to conventional parameter-based communication. %~\cite{DPSGD, ADPSGD}.  
%{\color{blue}should we mention our assymptions, \ie, roughly equal speed compute nodes?} \youjie{I don't think it as very important, because all related works and us use the identical GPUs over all nodes. Also, in the Scaling Efficiency part, I take the standard assumption from Facebook saying that the same batchsize on each node. As a results, the same batchsize + the same GPU = the same computation time. I think readers in distributed learning field should take this as default. }

Due to local pipelined training, balanced communication, and compression, the communication time is no longer part of the critical path, \ie, it is completely masked by computation, leading to linear speedup of end-to-end training time as the cluster size scales. Finally, we prove the convergence of \OURS for convex and strongly convex objectives  by adjusting the proof of~\cite{langford2009slow, SSP}.% and show slightly faster convergence rate than~\cite{SSP}.

\subsection{Timing Models and Decentralized \OURS}
\label{sec:timing}
\textbf{Timing Model: }
We propose timing models based on decentralized synchronous SGD to analyze the wall-clock runtime of training. Each training iteration consists of three major stages: model update, gradient computation, and gradient communication. Classical synchronous SGD (\figref{fig:AllFrame} (b)) runs local iterations on workers sequentially, \ie, each update depends on the gradient from the previous iteration, \ie, the iteration dependency is $1$. %\youjie{I added this to match the following sentence ``...relaxes the iteration dependency to $K$...''}
Therefore the total runtime of  synchronous SGD can be formulated easily as:
\begin{equation}
	l_{\text{total\_sync}} = T \cdot (l_{\text{up}} + l_{\text{comp}} + l_{\text{comm}}),
\end{equation}
where $T$ denotes the total number of training iterations and $l_{\text{up}}, l_{\text{comp}}, l_{\text{comm}}$ refer to the time taken by update, compute, and communication, respectively. It is apparent that synchronous SGD depends on the sum of execution time taken by all stages, which leads to long end-to-end training time.

\begin{figure}[!t]
  	\centering
  	 \includegraphics[width=\linewidth]{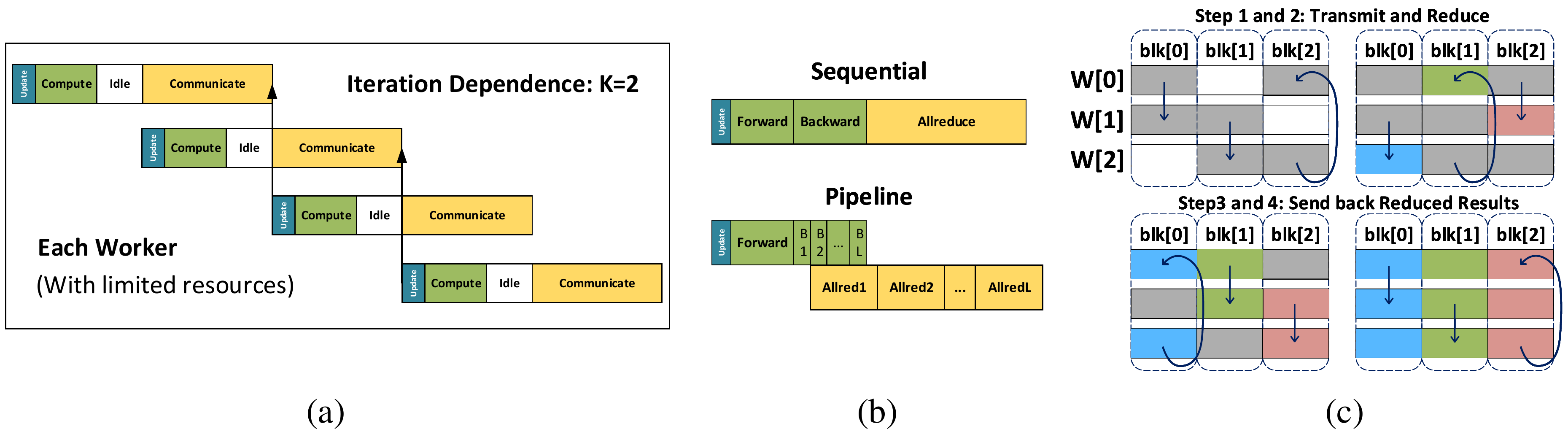}
   	\caption{Timing model of \OURS: (a) each worker with limited resources,  (b) sequential v.s. pipelined gradient communication, and (c) an example of gradient communication: Ring-AllReduce.}
	\label{fig:TimingModel}
	\vspace{-2.5ex}
\end{figure}

On the contrary,  \OURS relaxes the iteration dependency  to $K$, \ie, each update depends only on the gradients of the $K$-th last iteration. This enables  interleaving between neighboring iterations while maintaining globally synchronized communication, as shown in \figref{fig:AllFrame} (c). If we assume ideal conditions where both computation resources (CPU, GPU, other accelerators) and communication resources (communication links) are unlimited or abundant in counts/bandwidth, then the total runtime of  \OURS is:
\begin{equation}
	l_{\text{total\_pipe}} = T/K \cdot (l_{\text{up}} + l_{\text{comp}} + l_{\text{comm}}),
\end{equation}
where $K$ denotes the iteration dependency or the gradient staleness. We observe that the end-to-end training time in \OURS can be shortened by a factor of $K$. However, the ideal resource assumption doesn't hold in practice, because both computation and communication resources are strictly limited on each worker node in today's distributed systems. %, \ie each work has a small number of GPUs and tight bandwidth of communication links. 
As a result, the timing model for distributed learning is resource bound, either communication or computation bound, as shown in \figref{fig:TimingModel} (a), \ie, the total runtime is:
\begin{equation}
	l_{\text{total\_pipe}} = T \cdot \max(l_{\text{up}} + l_{\text{comp}}, l_{\text{comm}}),
	\label{eq:TotalPipe}
\end{equation}
where the total runtime is solely determined by either computation or communication resources, regardless of $K$ (when $K \geq 2$). Also, since gradient updates are always delayed by $(K-1)$ iterations, increasing  $K > 2$  only harms, \ie, the optimal value of $K = 2$ for \OURS with limited resources. Hence, the staleness of gradients is limited to $1$ iteration, \ie, the minimal staleness achievable in asynchronous updates. Besides,   we generally prefer a computation-bound setting for distributed training system, \ie, $l_{\text{up}} + l_{\text{comp}} > l_{\text{comm}}$. % the large communication overhead harms the scalability or speedup of large cluster, which will be discussed in 
To achieve this we discuss compression techniques in \secref{sec:compression}. 

%\youjie{Level two: gradient communication pipeline}
%\youjie{Level three: allreduce model}

In addition to  pipelined execution of iterations, we also analyze pipelined gradient communication within each iteration to reduce train time. %the communication latency.
Computation of gradients, \ie, the backward-pass, and communication of gradients are often  executed in a strictly sequential manner (see \figref{fig:TimingModel} (b)). However,  pipelined gradient communication, \ie, communicating gradients immediately after they are computed, is feasible. Again, we assume limited resources  and compare the sequential and pipelined gradient communication in~\figref{fig:TimingModel} (b).

%------------------------------------------------------------------------------
%\begin{figure}[!tbp]
%  \begin{subfigure}[b]{0.35\textwidth}
%    \includegraphics[width=\textwidth]{figures/ResourceLimited.pdf}
%    \caption{Each worker with limited resources.}
%    \label{fig:ResouceLimit}
%  \end{subfigure}
%\hfill
%  \begin{subfigure}[b]{0.25\textwidth}
%    \includegraphics[width=\textwidth]{figures/Figure_PipeComm.pdf}
%    \caption{Sequential vs. pipelined gradient communication.}
%    \label{fig:GradPipe}
%  \end{subfigure}
%\hfill
%  \begin{subfigure}[b]{0.35\textwidth}
%    \includegraphics[width=\textwidth]{figures/RingAllreduce.pdf}
%    \caption{An example of gradient communication: Ring Allreduce.}
%    \label{fig:RingAllreduce}
%  \end{subfigure}
%  
%   	\caption{Timing Model of \OURS. {\color{blue}can we align those figures a little more and maybe make them slightly bigger?}}
%	\label{fig:TimingModel}
%	\vspace{-2.5ex}
%\end{figure}

%------------------------------------------------------------------------------
To analyze the detailed timing of the those two approaches, we use the timing models for communication~\cite{ThakurRG05}. Communication of gradients is  an AllReduce operation which aggregates the gradient vector from all workers, performs the sum reduction element-wise, and then sends the result back to all. In practice, the underlying algorithms are much more involved~\cite{ThakurRG05}. For example, Ring-AllReduce,  one of the fastest  AllReduce algorithms, %performs circular-like reduction and transfer of multiple blocks of the gradient vector over the workers, as shown in~\figref{fig:TimingModel} (c).  
performs  gradient aggregation collectively among workers through balanced communication.  As shown in~\figref{fig:TimingModel} (c),  each worker transmits only a block of the entire gradient vector to its neighbor and performs the sum reduction on the received block. This ``transmit-and-reduce'' runs in parallel on all workers, until the gradient blocks are fully reduced on a worker (different for each block). Afterwards those fully reduced blocks are sent back to the remaining workers along the virtual ring. This approach optimally utilizes the network bandwidth of all nodes. %\youjie{rebuttal content No.1}

%\OURS employs a decentralized Allreduce algorithm with a flat hierarchy where workers perform the summation collectively through balanced communication. As shown in Figure~2(c) (see paper), each worker transmits only a block of the entire gradient vector to its neighbor and performs the sum reduction on the received block. This ``transmit-and-reduce'' runs in parallel on all workers, until the gradient blocks are fully reduced on a worker (different for each block). Afterwards those fully reduced blocks are sent back to the remaining workers along the virtual ring. This approach optimally utilizes the communication network which is often the real bottleneck.  

Adopting the Ring-AllReduce model of~\cite{ThakurRG05}, we obtain the total runtime of \OURS with sequential gradient communication under the limited resource assumption via:  
\begin{equation}
	l_{\text{total\_pipe\_s}} = T \cdot \max \left(l_{\text{up}} + l_{\text{for}} + l_{\text{back}}, \ 2(p-1)\cdot\alpha + 2(\frac{p-1}{p})\cdot n \cdot \beta + (\frac{p-1}{p})\cdot n\cdot \gamma + S \right),
	\label{eq:SeqGradComm}
\end{equation}
where  $l_{\text{for}}$ and $l_{\text{back}}$ denote forward-pass and backward-pass time,   $p$ denotes the number of workers, $\alpha$ the network latency, $n$ the model size in bytes, $\beta$ the byte transfer time, $\gamma$ the byte sum reduction time, and $S$ the global synchronization time. 
	
Similarly, we obtain the total runtime of \OURS with pipelined gradient communication via:
\begin{equation}
	l_{\text{total\_pipe\_p}} = T \cdot \max\left(l_{\text{up}} + l_{\text{for}} + l_{\text{b}}, \ 2(p-1)L\cdot\alpha + 2(\frac{p-1}{p})\cdot n \cdot \beta + (\frac{p-1}{p})\cdot n\cdot \gamma + L\cdot S \right),
	\label{eq:PipeGradComm}
\end{equation}
where $L$ denotes the number of gradient segments, and $l_{\text{b}}$ denotes the backward-pass time taken by the first segment.

Based on \equref{eq:SeqGradComm} and \equref{eq:PipeGradComm} we note: if a pipelined system remains communication bound, then sequential gradient communication is  preferred over the pipelined gradient communication (\equref{eq:SeqGradComm}  is smaller than \equref{eq:PipeGradComm} due to  positive $L$). %\youjie{because the second term in the $\max$ function will be chosen and  \equref{eq:SeqGradComm}  will always be smaller \equref{eq:PipeGradComm} due to the positive $L$. } 
In practice,  distributed training of large models is often communication bound,  making sequential exchange the best option. %{\color{blue}don't understand your reasoning here; can you provide more details?} %{\color{blue}so it's best in both?}

%% conventional sync sgd below
%Besides, the comparison between sequential and pipelined gradient communication also applies to the decentralized synchronous SGD in~\figref{fig:D-Sync}. We find that sequential approach is preferred if the following condition hold: {\color{blue}why is this important}
%\begin{equation}
%	l_{\text{back}} < 2(p-1)(L-1)\cdot\alpha + (L-1) \cdot S
%\end{equation}  
%In practice, this condition is easily satisfied in the task of distributed training of deep neural network using today's computation resources, where computation of backward-pass is much smaller than right-hand-side bound consisting of large $p$ and $L$. {{\color{blue}you need to explain where this condition comes from}}

To sum up,  based on our timing models, we find: \textbf{\OURS is optimal for $K = 2$, system is compute bound (after compression), and  sequential gradient communication is used}. Note that although our model is derived based on the Ring-AllReduce, this conclusion also applies to other AllReduce algorithms, such as recursive doubling, recursive halving and doubling, pairwise exchange, \etc ~\cite{ThakurRG05}.

\textbf{Decentralized Pipeline SGD:} %\label{sec:dec_pipe_sgd}
%-------------------------------------------------------------------------------------------------- 
\begin{algorithm}[!t]
\setstretch{1.1}
\fontsize{8.5}{9.0}\selectfont

\caption{Decentralized \OURS training algorithm for each worker.}
\label{alg:OurAlg}
\vspace{-1.5ex} 
\tline  
On the computation thread of each worker:
\tline
\vspace{2pt} 
\begin{algorithmic}[1]
   \STATE Initialize by the same model $w[0]$, learning rate $\gamma$, iteration dependency $K$, and number of iterations $T$. 
   \FOR{ $t = 1, \ldots, T$}
   	  \STATE Wait until aggregated gradient $g^{\text{c}}_{\text{sum}}$ in compressed format at iteration $[t-K]$ is ready
   	  \STATE Decompress gradient $g_{\text{sum}}[t-K] \leftarrow \text{Decompress}(g^{\text{c}}_{\text{sum}} [t-K])$
   	  \STATE Update $w[t] \leftarrow w[t-1] - \gamma \cdot g_{\text{sum}}[t-K] $
      \STATE Load a batch $\cB$ of training data
      \STATE Forward pass to compute current loss $f_\cB$
      \STATE Backward pass to compute gradient $g_{\text{local}}[t] \leftarrow \frac{\partial f_\cB}{\partial w[t]}$
      \STATE Compress gradient $g^{\text{c}}_{\text{local}}[t] \leftarrow \text{Compress}(g_{\text{local}} [t]) $
      \STATE Denote local gradient $g^{\text{c}}_{\text{local}}[t]$ as ready
   \ENDFOR
\end{algorithmic}
\vspace{2pt}
\tline 
On the communication thread of each worker:
\tline
\vspace{2pt} 
\begin{algorithmic}[1]
   \STATE Initialize aggregated gradients $g^{\text{c}}_{\text{sum}}$ of iteration $[1-K, 1-K+1, ..., 0]$ as zero and mark them as ready
   \FOR{ $t = 1, \ldots, T$}
	  \STATE Wait until local gradient $g^{\text{c}}_{\text{local}}[t]$ is ready
      \STATE AllReduce $g^{\text{c}}_{\text{sum}}[t] \leftarrow \sum g^{\text{c}}_{\text{local}}[t]$
      \STATE Denote aggregated gradient $g^{\text{c}}_{\text{sum}}[t]$ as ready
   \ENDFOR
\end{algorithmic}
\vspace{2pt} 
\bline
\vspace{-3ex} 

\end{algorithm}
%-------------------------------------------------------------------------------------------------- 
Guided by the timing models, we develop the decentralized \OURS framework illustrated in~\figref{fig:AllFrame} (c) where neighboring training iterations on workers are interleaved with a width of $K = 2$ while the execution within each iteration remains strictly sequential. Decentralized workers perform pipelined training in parallel with synchronization on gradient communication after every iteration. Due to the synchronous nature of our framework, the gradient update is always delayed by $K-1$ iterations, which enforces a deterministic rather than an uncontrolled staleness. In our optimal setting, the number of iterations for a delayed update is 1, as compared to $O(p)$ where $p$ is the cluster size in the conventional asynchronous  parameter server training~\cite{SSP, ADPSGD, TensorFlowOSDI}. Importantly, our framework still enjoys the advantage of an asynchronous approach -- interleaving of training iterations to reduce  end-to-end runtime. Also, different from the parameter server architecture, we don't congest the head node. Instead, in our case, every worker is only responsible for aggregating part of the gradients in a balanced manner such that communication and aggregate operation time are much more scalable. %\youjie{new} Although the synchronous SGD might also support balanced communication, it needs to swallow the entire execution latency especially from the slowest worker at each iteration. This becomes more sever as the cluster scales. In our \OURS, however, the drift of latency can be tolerated or completely hidden by the local interleaving, making our system more scalable in practice.

%{\color{blue}we should mention the disadvantage if the compute nodes are not equally fast; this is an assumption which is part of our model} 

More formally, we outline the algorithmic structure of our implementation for each worker in~\algref{alg:OurAlg}. To be specific, each worker has two threads: one for computation and one for communication, where the former thread consumes the aggregated gradient of the $K$-th  last iteration and generates the local gradient to be communicated, and the latter thread exchanges the local gradient and buffers the aggregated results to be consumed by the former thread.

\subsection{Compression in \OURS}
\label{sec:compression}
To further reduce the communication time we  integrate lossy compression into our decentralized \OURS framework. %  as an important component~\algref{alg:OurAlg}, not only because the communication overhead always dominates the runtime due to larger model size, larger cluster size, and slow improvement in network technology, but also because our decentralized framework offers maximized opportunity for lossy compression on communication data. 
Unlike the conventional parameter server or recent decentralized framework transferring parameters over the network~\cite{Dean2012,Chilimbi2014,MuLiOSDI2014,MuLiNIPS2014,MoritzICLR2016,IandolaCVPR2016, SSP, ADPSGD, DPSGD}, our approach communicates only gradients and we justified empirically that gradients are much more tolerant to lossy compression than the model parameters. % during training process.     
This seems intuitive since reducing the precision of parameters in every iteration harms the final precision of the trained model directly.

Importantly, as  mentioned in~\secref{sec:timing}, compressing the communication overhead contributes to the optimal setting of  \OURS. % since the original communication bound system can turn into be computation bound with a shorter communication latency. 
Once  \OURS  is completely computation bound,  linear speedups of end-to-end training time can be realized as the cluster size increases. Analytically, we show this observation by deriving the scaling efficiency using the timing model given in \equref{eq:TotalPipe}. Assume that: 1) the singe-node training takes $T_{\text{single}}$ iterations to complete with an execution time of $l_{\text{single}}$ taken by each iteration; 2) given a \OURS cluster with $p$ workers we use the same batch size on each worker as the single-node~\cite{Facebook1Hour}; 3) the single node and \OURS train the same epochs on the dataset.  From 2) and 3), we find that the total number of iterations required for \OURS is $T_{\text{single}}/p$, because \OURS has a $p$ times larger batch size while still training the same number of samples. From this we obtain the scaling efficiency $SE$ of \OURS via
\begin{equation}
	\operatorname{SE} = \frac{\operatorname{Actual\ Speedup}}{\operatorname{Ideal\ Speedup}} = \frac{ \frac{l_{\text{single}} \cdot T_{\text{single}}}{ l_{\text{total\_pipe}} } } {p} = \frac{ \frac{l_{\text{single}} \cdot T_{\text{single}}}{ \max(l_{\text{up}} + l_{\text{comp}},\ l_{\text{comm}}) \cdot \frac{ T_{\text{single}} }{p}  } } {p} = \frac{l_{\text{up}} + l_{\text{comp}}}{\max(l_{\text{up}} + l_{\text{comp}},\ l_{\text{comm}})}. 
\end{equation}
Thus, we showed that once our system becomes compute bound with compressed communication,  \OURS  can achieve linear speedup as the cluster scales, \ie, $SE = 1$.  

%{\color{blue}shouldn't we use Eq. (4) rather than Eq. (3) here?} \youjie{We already use the Eq.(4). } {\color{blue}but you divide by $p$ for $l_{total_pipe}$ which you shouldn't. What am I missing?} \youjie{Because we are training the same epoches on the dataset and \OURS is $p\times$ larger in batchsize, making the total number of iteration divided by $p$. Added assumption 3).} {\color{blue}I  think you should add a sentence to describe this more}

%{\color{blue}can you improve the writing of the following paragraph; it's hard to understand;}

To maintain applicability of Ring-AllReduce, we choose two simple compression approaches: truncation and scalar quantization. Truncation drops the less significant mantissa bits of floating-point values for each gradient. The scalar quantization discretizes each gradient value into an integer of limited bits, with a quantization range determined by the maximal element of a gradient vector. % where target gradient resides. 
Due to their simplicity, we easily parallelize those compression approaches to minimize overhead. 
%
%Note that the compression itself can be compute-heavy, increasing the total runtime. 

Note that  compression itself can be compute-heavy %(\eg, complex ones often require sorting over hundreds of MB of gradients or even sequential execution) 
and the introduced computation overhead can outweigh the benefit of compressed communication. 
Particularly when considering that AllReduce based communication performs multiple steps to transfer and reduce the data (see \figref{fig:TimingModel} (c)), requiring  repeated invocation of compression and decompression, \ie, for each ``transmit-and-reduce'' step, with an invocation complexity linear in cluster size. 
Therefore, many proposed complex compression techniques~\cite{Microsoft_1BitSGD, Amazon_FixThresh, Lawrence_Adaptive, IBM_AdaComp, TernGradNIPS2017, deep_g_compression} often fail in the communication-optimal AllReduce setting, resulting in longer wallclock time.  
For these reasons,  compression embedded inside AllReduce must be light, fast and easy to parallelize, such as a floating-point truncation or our element-wise quantization. 
%\youjie{rebuttal content No.2}

% Compression with pipelined Allreduce
\begin{figure}[!t]
  	\centering
  	 \includegraphics[width=0.8\linewidth]{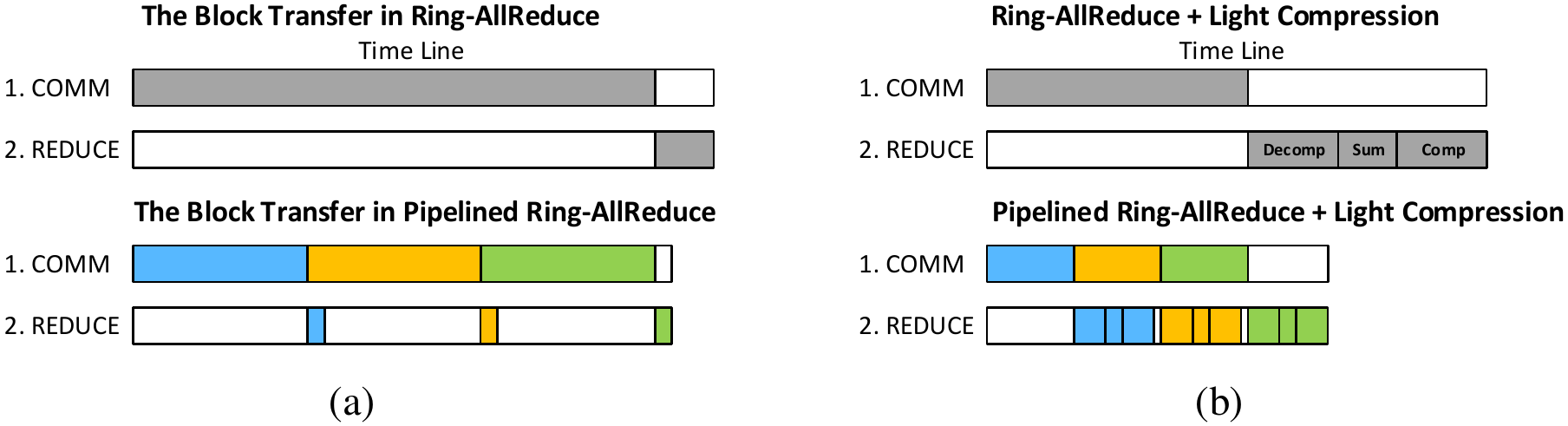}
   	\caption{Pipelining within AllReduce: (a) block transfer in native Ring-AllReduce and pipelined Ring-AllReduce, and (b) block transfer with light-weight compression.}
	\label{fig:PipeAllReduce}
	\vspace{-2.5ex}
\end{figure}

Indeed,  \textit{pipelining within AllReduce} can help alleviate the heavy overhead of complex compression. However, its benefit might still be limited. Instead of pipelining of training iterations as in \OURS, \textit{pipelining within AllReduce} interleaves the gradient communication and reduction within each AllReduce process, as illustrated in~\figref{fig:PipeAllReduce} (a). Since the communication time is often larger than the reduction time, the latter can be hidden by the former. Once compression is used (as in \figref{fig:PipeAllReduce} (b)), the two stage pipeline becomes (decompression, sum, compression) and (compressed communication) such that light compression overhead can be masked completely. Although complex compression may also benefit from the pipelined AllReduce, the improvement is limited because the time spent by complex compression often outweighs the communication time. For example, we implemented~\cite{TernGradNIPS2017} within the pipelined AllReduce and found that the compression overhead is $1.6\sim2.3\times$ the uncompressed communication time and $25.6\sim36.8\times$ the compressed communication time for the benchmarks in~\secref{sec:exp}, in which case the heavy overhead cannot be masked. Complete masking requires the compression overhead to be smaller than the compressed communication. In the remainder, we only consider light compressions (truncation/quantization) with native AllReduce.

%\youjie{rebuttal content No.3. I think it is better to put pipelined Allreduce here, because the major merit of having pipelined Allreduce is to alleviate compression overhead.}
%\youjie{Not sure if we should put experimental data here or explicitly attack other's work.}

%
%For this reason, many proposed compression algorithms~\cite{Microsoft_1BitSGD, Amazon_FixThresh, Lawrence_Adaptive, IBM_AdaComp, TernGradNIPS2017, deep_g_compression} are not applicable to Allreduce based communication if evaluated on wall-clock time, which seems to be a hardly known fact. %, that is crucial for 

\subsection{Convergence}
To prove the convergence of \OURS we adapt the derivation from parameter-server based asynchronous training~\cite{SSP, langford2009slow}. We can show that the convergence rate of \OURS for convex objectives via SGD is $8FL\sqrt{\frac{K}{T}}$, where $K=2$, $F$ and $L$ are constants for gradient distance and Lipschitz continuity, respectively. We can also show  the convergence of \OURS for strongly convex functions, and find a rate of $O(\frac{\log T}{T})$ for gradient descent. These rates are consistent with~\cite{SSP, langford2009slow}. Due to the page limit we defer details to the supplementary material. %We leave the proof details to appendix due to limited space
\section{Experimental Evaluation}
\label{sec:exp}

In this section, we demonstrate the efficacy of our approach on four benchmarks using three datasets: MNIST~\cite{LeCunIEEE1998}, CIFAR100~\cite{Krizhevsky2009} and ImageNet~\cite{RussakovskyIJCV2015}. We  briefly review  characteristics of those datasets before discussing  metrics and setup, and finally presenting  experimental results and analysis.

\textbf{Datasets and Deep Net Architecture}
\begin{itemize}[leftmargin=*]
	\vspace{-1ex}
	\item \texttt{MNIST:} 
				The MNIST dataset %, which  is a subset of a larger set available from NIST,
				%is a benchmark for classification of handwritten digits, 
				consists of 60,000 training and 10,000 test images, each showing one of ten possible digits. The images  
				%The size of a digit image is normalized to 
				 are of size  $28\times 28$ pixels with
				 digits located at the center of the images. %The predictor $F(x,w)$ aims at retrieving the correct digit category given the raw input image $x$. 
				We use a classical 3-layer perceptron, MNIST-MLP, with both hidden layers being 500-dimensional and with a global batch size of $100$.
				%It has 10 classes (\ie, digit 0,1,2,...9).
				%reference: http://yann.lecun.com/exdb/mnist/
				
	\item \texttt{ImageNet:} 
				%The most recent ImageNet is made up of 14,197,122 examples in total.
				For our experiments we use 1,281,167 training and 50,000 validation examples from the ImageNet challenge. 
				Each example comprises a color image of 256$\times$256 pixels and belongs to one of  1000 classes. 
				We use the classical AlexNet~\cite{KrizhevskyNIPS2012} and ResNet~\cite{HeCVPR2016}, both with a global batch size of $256$.
				%reference: http://www.image-net.org/
				%http://www.image-net.org/papers/imagenet_cvpr09.pdf
				
	\item \texttt{CIFAR100:} 
				The CIFAR100 dataset is composed of 50,000 training and 10,000 test examples with 100 classes. %Specifically, it has 20 superclasses (\ie, aquatic mammals, fish, flowers, \etc), each of which is further divided into 5 classes (\eg, beaver, dolphin, otter, seal, and whale for aquatic mammals). 
				%Each class contains 600 images. %(\ie, 500 training and 100 testing images, respectively). %The prediction task is 100 class classification. 
				%The quick CIFAR100 architecture %\new{from a Chainer tutorial~\cite{CIFAR100Chainer}} is used for benchmarking this datasets.
				The simple AlexNet-style CIFAR100 architecture in~\cite{RenjieCNN} is used for benchmarking this datasets.
				It consists of 3 convolutional layers and 2 fully connected layers followed by a softmax layer. %\new{This small model is trained with a global batch size of $100$.}
				The detailed parameters are available in~\cite{RenjieCNN}.
				Importantly, we adapt this 5 layer CIFAR100-CNN into a convex optimization benchmark, CIFAR100-Convex, to match our proof of convergence. The convexity is achieved by training only the last fully connected layer while fixing the parameters of all previous layers. %In practice, we divide the original 5-layer model into two: the front net with the front 4 layers, and convex net with the last layer and softmax, where the former only performs the forward pass based on pretrained parameters and the latter trains with cross-entropy loss from randomly initialized model. In experiments, we only measure the performance of the convex net. 
				%reference: https://www.cs.toronto.edu/~kriz/cifar.html
	%\vspace{-1ex}
\end{itemize}

%\textbf{CIFAR-10:} 
%The CIFAR-10 dataset %, which is a benchmark for classification of images, 
%is composed of %60000 images in total, where 
%50,000 training and 10,000 test examples. 
%Each example is a color image of 32$\times$32 pixels. It has 10 classes (\ie, airplane, automobile, bird, \etc) with 6000 images per class. The prediction task is again 10 class classification. The quick CIFAR-10 architecture of Caffe~\cite{JiaCaffe2014} is used for benchmarking this datasets. It consists of 3 convolutional layers and 1 fully connected layer followed by a softmax layer. The detailed parameters are publicly available on the Caffe~\cite{JiaCaffe2014} website.
%%reference: https://www.cs.toronto.edu/~kriz/cifar.html

%------------------------------------------------------------------------------
\begin{figure}[!t]
	\centering
	\begin{center}
	\includegraphics[max size={\textwidth}{\textheight}]{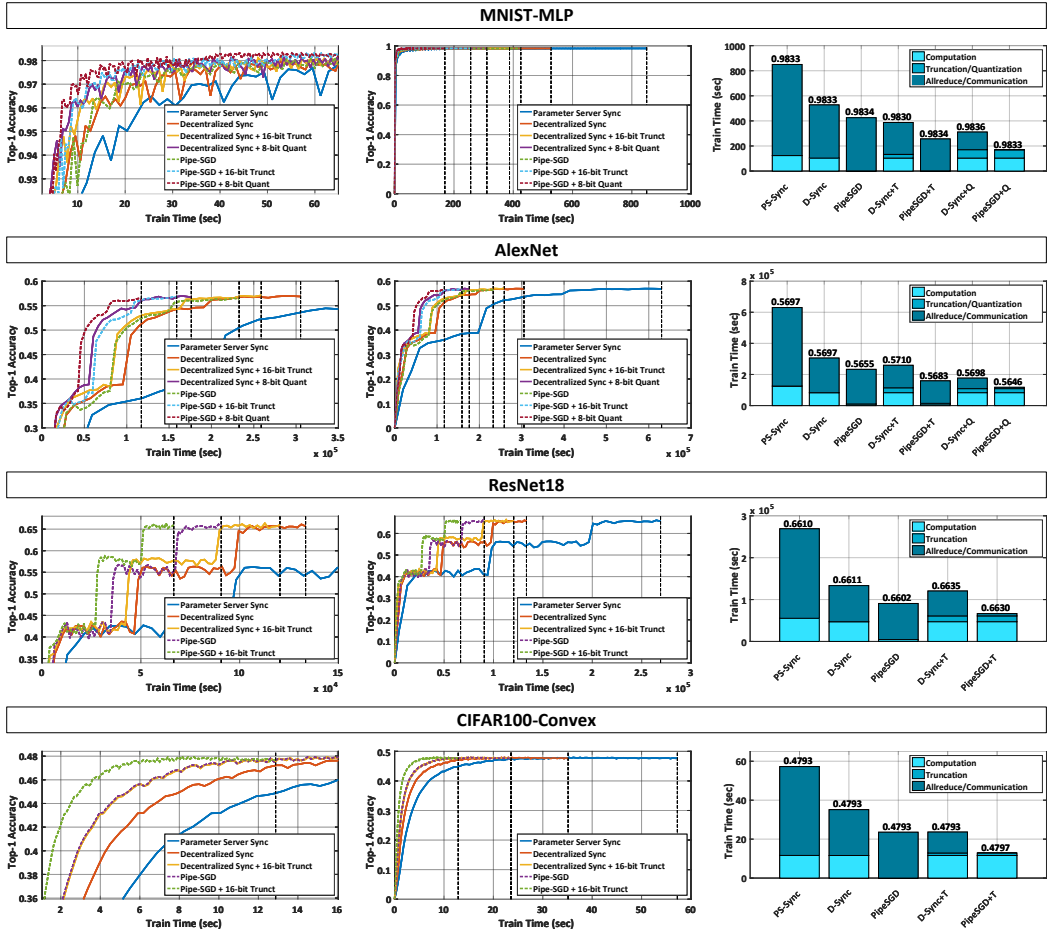}
	\end{center}
	\caption{\textbf{Experimental results:} Each row shows different benchmarks. The left two columns show convergence via test/validation accuracy \vs wallclock training time, where the first column is an inset of the second one. The right most column shows the detailed timing breakdown of end-to-end training. Note that the final top-1 accuracies on test/validation set are labeled on top of the bars. }
	\label{fig:AllFig}
	\vspace{-3.0ex}
\end{figure}
%------------------------------------------------------------------------------

\textbf{Metrics and Setup}

We measure the wall-clock time of end-to-end training, \ie, the same number of iterations for different settings. For each benchmark, we evaluate the timing model we proposed using end-to-end train time and detailed timing breakdowns. We plot the test/validation accuracy over training time to evaluate the actual convergence. Also, final top-1 accuracies on the test/validation set are reported. %\alex{test set or validation set? also check caption of figure}\youjie{we measure all accuracies on test set for sure in our LSDN. we don't have validation set in LSDN and use test set for accuracy evaluation.}
%\alex{this doesn't seem to make sense since the test set labels are not public: https://stackoverflow.com/questions/40744700/how-can-i-find-imagenet-data-labels; what data did you evaluate on? validation set?}
For the setup, we use a cluster of four nodes, each of which consists of a Titan XP GPU~\cite{TitanXP} and a Xeon CPU E5-2640~\cite{IntelXeon}. We employ an additional node as the parameter server to support the conventional centralized design. All nodes are connected by 10Gb Ethernet. 
%
%\new{Although we wish to scale up our cluster, as an academic group we don't have access to more 10GbE NICs, GPUs, and machines.} \youjie{another rebuttal content attacks our scalability. Should we bring this sentence?}
%, to evaluate the scalability of \OURS empirically without interference. Analytically, we  showed  scalability of \OURS   in Eq.~(7) and we are sure that  assumptions hold  in a large scale setup.} 
%
We implement a distributed training framework in C++ using CUDA 8.0~\cite{CUDA}, MKL 2018~\cite{MKL}, and OpenMPI 2.0~\cite{OpenMPI}, which supports the parameter-server and \OURS approach. %Note that \OURS can also be implemented in Caffe~\cite{JiaCaffe2014} with MPI.

%Note that the state-of-the-art network architectures of datacenter at large Internet companies such as Google and Facebook use 1$\sim$10Gbps network connections within a rack and 10$\sim$100Gbps connections for the oversubscribed links between the top of rack switches~\cite{facebookcenter, 43837}. As the clusters running the training applications are connected to the top of rack switches, we did not consider supporting 40$\sim$100Gbps network connections for our experiments.
%Note that \OURS can also be implemented in Caffe~\cite{JiaCaffe2014} with MPI. However, our custom distributed execution framework is more amenable for integration with asynchrouns or pipeline training as well as lossy compressions. In our custom training framework, all the computation steps of DNN training such as forward and backward propagations are performed on the GPU (also CPU compatible) while communication is handled via \textsf{\small OpenMPI} APIs. Besides, our framework implements diverse distributed training architectures and communication algorithms using various types of \textsf{\small OpenMPI} APIs to exchange gradients and weights. 

%We implement the \OURS in C++ using NVIDIA CUDA GPU acceleration and  Intel MKL (Math Kernel Library) to optimize the compute performance. For communication between workers we take advantage of OpenMPI and a cluster comprised of four compute nodes, each of which is equipped with an NVIDIA Titan GPU, an Intel Xeon CPU and a 10Gb Ethernet Network Interface Card (NIC).
%

\textbf{Results and Analysis}

We evaluate the performance of three different frameworks: parameter server with synchronous SGD (PS-Sync), decentralized synchronous SGD (D-Sync), and \OURS. Our compression schemes, \ie, 16-bit truncation (T) and 8-bit quantization (Q), are also applied to AllReduce communication in D-Sync and \OURS. %\alex{we never explain quantization, do we?}\youjie{we do mention the quantization in the Compression section. we only mentioned it slightly to avoid contribution conflict with our PCA compression paper from USC side.} 
Evaluation results are summarized in \figref{fig:AllFig} where the first two columns show the convergence performances and the third column shows detailed timing breakdowns with final accuracies labeled. 

\textit{Convergence}: From \figref{fig:AllFig}, we observe: decentralized approaches, \ie, D-Sync and \OURS, converge much faster than the parameter server even without compression, and \OURS shows the fastest convergence among these frameworks, especially when compression is applied.  For example, the convergence curve of the CIFAR100-Convex shows that D-Sync is around $40\%$ faster than PS-Sync and \OURS is another $37\%$ faster than  D-Sync. The advantage of \OURS is further boosted by  compression, \ie, truncation in this case, and demonstrates an additional $46\%$ faster convergence than the D-Sync with the same compression scheme. Therefore  \OURS prevails with a great margin.

\textit{Timing Breakdown}: %From the third column of ~\figref{fig:AllFig}, %, we note that distributed training systems are mostly communication bound and thus justifies our assumption in the~\secref{sec:timing}. Also, 
From \figref{fig:AllFig}, the comparison between centralized and decentralized designs shows $50\%$ reduction in uncompressed communication time, thus justifying the efficacy of balanced communication. Once compression is applied, further reduction is observed. However, the actual improvement in D-Sync is not ideal considering  compression factors of $2\times$ for truncation and $4\times$ for quantization, because the compression overhead is paid at the critical path of D-Sync. In contrast, our \OURS can hide this overhead together with computation due to the pipelined nature, as shown in ``D-Sync+T'' \vs ``PipeSGD +T'' in the MNIST benchmark. As communication is further reduced by quantization, the system becomes compute bound and \OURS switches to hide the communication instead, thus reaching the optimal setting of \OURS. This optimum can also be achieved via the simplest truncation for models with less dominant communication time, \eg, ResNet18 and CIFAR100-Convex. As a result, our approach achieves a speedup of $2.0\sim3.2\times$ compared to D-Sync and $4.0\sim5.4\times$ compared to PS-Sync for these benchmarks. Note that these speedups are based on the comparison between different approaches in the same cluster without scaling the cluster size.

%  \alex{Why larger than $5$ is possible?}\youjie{Since we are comparing our approach to parameter server approach, both based on the same cluster. We are not comparing 4 nodes v.s. 1 node. We only evaluate different approachs on the same cluster size for fairness. (Well, the parameter server has 5 nodes. doesn't matter.) } \alex{we need to explain this, people will ask why the number is larger than 5?}

\textit{Accuracy}: Considering the potential drawback of the 1-iteration staled update and lossy compression in \OURS, we also evaluate the final test/validation accuracies after end-to-end training, as shown in~\figref{fig:AllFig}. Interestingly, in our optimal settings ``PipeSGD +T/Q,'' we find that only AlexNet drops top-1 accuracy by $0.005$  compared to baseline D-Sync while all other benchmarks show slightly improved accuracies. %\youjie{warm up disclosure here} 
To obtain the best accuracies for the two large non-convex models such as AlexNet and ResNet, we employ a similiar warm-up scheme as in~\cite{deep_g_compression}, \ie, we don't turn on the pipelined training until the $5$-th epoch, before which we still stick to D-Sync training to avoid the undesirable gradient change in the initial stage. Since the warm-up period is marginal compared to total number of epochs, the system performance benefits from \OURS most of the time. Note that for smaller models, especially convex ones (\eg, CIFAR100-Convex), no warm-up is required.  
\section{Related Work}
%\vspace{-1ex}
\label{sec:related}
%Beyond already cited work we note that two approaches are related to \OURS. \youjie{Revisit this part soon}

Li \etal~\cite{MuLiOSDI2014,MuLiNIPS2014} proposed a parameter server framework for distributed learning and a few approaches to reduce the cost of communication among compute nodes, such as exchanging only nonzero parameter values, local caching of index list, and random skip of messages to be transmitted. Abadi \etal~\cite{TensorFlowOSDI} also proposed a centralized framework, TensorFlow, which incorporates model and data parallelism for training deep nets. Both works support the asynchronous setting to improve communication efficiency but without controlling the staleness of the gradient update. Ho \etal~\cite{SSP} proposed  SSP, another centralized asynchronous framework but with bounded staleness for gradients.  The key idea of SSP: 1) each worker has its own iteration index, 2) the slowest and fastest worker must be within $S$ iterations, otherwise, the fastest worker is forced to wait until the slowest worker catches up. %{\color{blue}don't understand whatever follows after this comment in this paragraph} \youjie{For example, the four workers in the \figref{fig:AllFrame} (a) are at the same iteration so the iteration drift is $0$ and thus within the $S$ bound. All workers are allowed to commit their gradient update to the centralized server asynchronously, so we see the $W1\sim W5$. It it evident that the slowest gradient push from worker3 is already staled by 3 iterations at the moment of update, which is $O(CluserSize - 1)$. }. 
However, this bound $S$ applies to the iteration drift among workers instead of directly on the stale updates of the parameter server. As a result, each worker within the bound can still commit their updates to the server asynchronously, making the last gradient update staled heavily. In the worst case, the staleness is linear in the cluster size. 

Lin~\etal~\cite{deep_g_compression} employed  AllReduce as the gradient aggregation method in their synchronous framework, but little is reported regarding  wallclock time benefits, especially considering that the full synchronous design suffers from the longest execution time among all workers. Besides, Lian~\etal proposed AD-PSGD~\cite{ADPSGD} which parallelizes the SGD process over decentralized workers in a completely asynchronous fashion. Workers run completely independently, and only communicate with a set of neighboring nodes to exchange trained weights, \ie, neighboring models are averaged to replace each worker's local model in each iteration. However, this approach suffers from uncontrolled staleness, which in practice increases with cluster size and the time taken by each iteration. In addition, such a communication method requires  each worker to act as the center node of a local graph, which  results in a local communication bottleneck. As a result, each worker suffers from long iteration time which further increases the staleness of weight updates. Although Lian~\etal~\cite{ADPSGD} compared their framework with the full synchronous design in wall-clock time, the performance turns out to be similar when  network speeds are roughly equal. % in the cluster. 

Recently, independent work~\cite{PipeDream} also proposed a distributed pipelined system for DNN training. Different from~\OURS, \cite{PipeDream} focuses on pipelining with model parallelism, partitioning the DNN layers onto different machines and pipelining the execution of the machines by injecting consecutive mini-batches into the first one. This approach reduces communication load since only activations and gradients of a subset of layers are communicated between machines. However, complex mechanisms (such as profiling, partitioning algorithm, and replicated stages) are necessary to balance the workload among different machines, otherwise compute resources turn idle. Furthermore, \cite{PipeDream} may suffer from staleness of the weight update, which is linear in the number of stages. This limits the effectiveness of model pipelining and throttles speedups.
\vspace{-0.2cm}
\section{Conclusion}
\label{sec:conclusion}
\vspace{-0.2cm}
We developed a rigorous timing model for distributed deep net training which takes into account network latency, model size, byte transfer time, \etc. Based on our timing model and realistic resource assumptions, \eg, limited network bandwidth, we assessed scalability and developed \OURS, a pipelined  training framework which is able to mask the faster of computation or communication time. We showed efficacy of the proposed method on a four-node GPU cluster connected with 10Gb links. Rigorously assessing wall-clock time for \OURS, we are able to achieve improvements of up to $5.4\times$ compared to conventional approaches.

% 9 pages for contents
%%%%%%% -- PAPER CONTENT ENDS -- %%%%%%%%

%%%%%%%%% -- BIB STYLE AND FILE -- %%%%%%%%
\clearpage % Unlimited Pages
\section*{Acknowledgement}
This work is supported in part by grants from NSF (IIS 17-18221, CNS 17-05047, CNS 15-57244, CCF-1763673 and CCF-1703575). 
This work is also supported by 3M and the IBM-ILLINOIS Center for Cognitive Computing Systems Research (C3SR). %- a research collaboration as part of the IBM AI Horizons Network. 
Besides, this material is based in part upon work supported by Defense Advanced Research Projects Agency (DARPA) under Contract No. HR001117C0053. The views, opinions, and/or findings expressed are those of the author(s) and should not be interpreted as representing the official views or policies of the Department of Defense or the U.S. Government.
\bibliographystyle{plain}
{\small
\bibliography{pipesgd}}
%%%%%%%%%%%%%%%%%%%%%%%%%%%%%%%%

%\clearpage
%\input{appendix}

\end{document}